\documentclass{article}

\usepackage{graphicx}
    \usepackage[final]{neurips_2023}

\usepackage[utf8]{inputenc} % allow utf-8 input
\usepackage[T1]{fontenc}    % use 8-bit T1 fonts
\usepackage{hyperref}       % hyperlinks
\usepackage{url}            % simple URL typesetting
\usepackage{booktabs}       % professional-quality tables
\usepackage{amsfonts}       % blackboard math symbols
\usepackage{nicefrac}       % compact symbols for 1/2, etc.
\usepackage{microtype}      % microtypography
\usepackage{xcolor}         % colors

\title{Combating the "Sameness" in AI Art: Reflections on the Interactive AI Installation \emph{Fencing Hallucination}}

\author{
  Weihao Qiu\thanks{Affiliated with Experimental Visualization Lab at UCSB MAT under the supervision of Prof. George Legrady. Weihao Qiu's personal website is www.q-wh.com} \\
  Media Arts and Technology Program\\
  University of California, Santa Barbara\\
  Santa Barbara, CA 93106 \\
  \texttt{wqiu@ucsb.edu} \\
  \And
  George Legrady \\
  Media Arts and Technology Program \\
  University of California, Santa Barbara\\
  Santa Barbara, CA 93106 \\
  \texttt{glegrady@ucsb.edu} \\
}

\begin{document}

\maketitle

\begin{abstract}
The article summarizes three types of "sameness" issues in Artificial Intelligence(AI) art, each occurring at different stages of development in AI image creation tools. Through the Fencing Hallucination project, the article reflects on the design of AI art production in alleviating the sense of uniformity, maintaining the uniqueness of images from an AI image synthesizer, and enhancing the connection between the artworks and the audience. This paper endeavors to stimulate the creation of distinctive AI art by recounting the efforts and insights derived from the Fencing Hallucination project, all dedicated to addressing the issue of "sameness".

\end{abstract}

\section{"Sameness" in AI Art}
% three short names
% depth -> improved another word
%

The "Sameness" in AI art refers to a recurring phenomenon in the field of AI-generated art, wherein artworks exhibit a pronounced uniformity or lack of distinctiveness. This uniformity may manifest as the repetition of established artistic styles, a deficiency in original creative expression, superficial imitation of artistic features without genuine understanding, or a limited range of variation in generated artworks. The concept underscores the challenge of achieving uniqueness in AI-generated art, prompting efforts to diversify such creations from established artistic conventions.

Three types of "sameness" issues have emerged along with the ongoing development of AI art tools. The first is caused by the lack of novel dataset. It started with experiments on the visual imitation of established artworks [1][2][3]. Except for a few cases[19], creating an original personalized image dataset for AI training has been beyond an individual's effort, most users of AI creation tools use off-the-shelf general-purposed datasets, consequently resulting in similar aesthetics of different AI artworks. [see sup. material]

The second type of "sameness" is a result of the lack of control in the creation process. Because replicating the style in the image dataset provides little visual novelty, other creative minds have chosen to break away from this practice. Through methods such as mixing different images in the training[17], latent sampling[7], and under-training, they have created images of visual indeterminacy[4]. These images consist of broken, chaotic, checker-like patterns and unnamed forms, which would be frowned on by the tool's creator but exploited by artists as an expressive space. Opposed to the designed function of the generative AI models, this method has provided users little control in the creation process, limiting range of the variations. [see sup. material] 

% informed decision-making and stopping the audience's aesthetic experience from extending beyond the superficial level. 

With diffusion models[5] being the latest breakthrough in AI image synthesis, artists can easily create various high-fidelity images[see sup. material]. The lowered technological barrier of these new tools greatly expands the users beyond creative programmers and AI researchers to everyone who wants to create. AI's incredible replicating capability has become the new normal, driving audiences to seek connection and depth in these AI artworks. With the text prompt being the artist's only input to the software, the audience has too limited information to understand the artist's decision-making and production process, stopping the aesthetic experience from extending beyond the superficial level. This reduces the perceived uniqueness of the produced art. Therefore, many artists have criticized these AI tools for showing "banal overall "sameness" across generated art." [6]

\section{Fencing Hallucination}

Fencing Hallucination is a Human-AI Interactive Installation[16] that emulates chronophotography using motion capture, motion synthesis, and pose-conditioned AI image synthesis [see sup. material]. The installation consists of two large displays for real-time interaction and final images, set up back-to-back. An audience starts by moving their body to interact with a virtual AI Fencer on the screen. As the audience moves actively, they get exciting responses from the AI fencer and trigger strobe flashes. At each flash, the other screen refreshed to a new chronophotograph updated by the fencing poses, making the final image increasingly complex as the audience extends their interaction time.

According to their feedback, audiences perceived the images from Fencing Hallucination more as real photographs taken during a unique in-person experience rather than images indifferently created by AI. People also showed a solid connection to the result images; they downloaded them on their phones to share with others later. Some people appreciated the photo so much that they returned to the installation to produce a more exciting picture. This observation indicates that Fencing Hallucination made progress in combating the "sameness" in AI art.

\section{Reflections}
The evaluation of the design of Fencing Hallucination to combat AI "sameness" is four-fold: 

\emph{1. Customized thematic dataset:} Fencing Hallucination trained specialized motion and image synthesis models using carefully collected datasets of fencing content.[see sup. material] This offered a thematic consistency throughout the experience, which presented AI beyond its generic capabilities.

\emph{2. Integrating full-body interactivity:} Unlike other diffusion-based AI artworks using text as the only input, Fencing Hallucination integrated body poses as the other predominant condition for its outcome. The poses-to-image translation[18] enabled full-body interactivity to enhance user engagement, strengthening the connection between the audience and the produced art. 
% Fencing Hallucination uses full-body interactivity to motivate user engagement in AI art production. The physical engagement strengthens the bond between the audience and the produced art. Compared with the AI art made with limited modality, participants appreciate more the art created from their unique full-body interactive experience. 
% The interactivity is implemented with a lightweight real-time motion synthesis model, which provides the motion data for Stable Diffusion to generate images.

\emph{3. Increased control of visual aesthetics by modular design:} Fencing Hallucination revived chronophotography, necessitating precise controls in the creative process. These controls were enabled by a modular workflow, with individual modules in charge of sub-tasks, including motion synthesis, pose-to-image translation, and post-processing for multi-exposure styles, each contributing to different aspects of the final image.

% reshaped AI art aesthetics by resurrecting
% rather than exclusively pursuing entirely novel ones
% Additionally, it produced images with detailed results achieved by overlaying multiple images, inspired by multi-exposure photography.

% Fencing Hallucination re-examines the assumed aesthetics in AI art, which values extensively on the search for novel visual styles yet to be found. Given the current AI's nature is to mimic the styles embedded in the dataset rather than to derive new ones, Fencing Hallucination explores alternatively the recycling potential of existing visual aesthetics, particularly chronophotography. The familiarity with visual aesthetics redirects the audience's attention from expecting visual spectacles to seeking depth and connection to the artwork, which are fulfilled through careful interactive experience design. On the other hand, Fencing Hallucination also amazed the audience by the incredible details in the result images. The single-image AI synthesis cannot produce sophisticated patterns shaped by body locomotion. Inspired by multi-exposure photography, Fencing Hallucination overcome this challenge by creating multiple images and overlay them into the final image.

\emph{4. Chose themes and narratives thoughtfully to stimulate tech-related philosophical reflections:} The choice to resurrect chronophotography aimed to challenge the assumption about AI art in its narrow search for unseen styles[8]. It also connects two technologies spanning 140 years, photography and AI image synthesis, which transformed visual art's realm. Through juxtaposing authentic photographic mimicry with the lack of camera usage, Fencing Hallucination prompts the audiences' reflections on technological evolution. By realizing human-AI co-creation of visual art, this work also dispels the tense, adversarial point of view on AI and alleviates the human-replacement concern on AIs. 

% Generative AI is performing transforming power to the field of visual art, deconstructing the ontology of a photography and blurring boundary between the real and the forged. The last technology that has similar transforming power in visual art is photography. Fencing Hallucination connects two technology spanning over 140 years. Juxtaposing the authenticity of the produced images with the fact that they were created purely with AI and positional data only, relying on no camera, the project invites the audience to reflect on the evolution of technology. Through the realization of human-AI co-creation of visual art, this work also dispels the tense, adversarial point of view on AI and alleviates the human-replacement concern on AIs. 

% \section{Conclusion}
% This article analyze the issue of "sameness" in AI Art and share the reflection on the project Fencing Hallucination which aims to combat this issue in three dimensions of aesthetics, technology, and socials. 

\section{Ethical Implication}
This paper described strategies for addressing the issue of sameness in AI art. One involves resurrecting an existing visual style, challenging the prevailing practice of using AI solely to seek entirely new visual styles. While this particular mimicry is rooted in a deep understanding of photographic culture and technology, it must be performed cautiously to avoid appearing as encouraging the mere superficial copying of established artists' works. Additionally, while the training data in this article was collected ethically from fencing game videos using motion estimation methods without copyright infringement concerns, the video scraping process may have inadvertently included non-public footage. Ensuring the anonymity and security of the data is a vital consideration for all artists using self-collected datasets.

% Generative AI raises concerns about the ownership and authorship of creative works. As AI systems generate art, literature, and music autonomously, questions emerge regarding who should be recognized as the rightful creator or owner of the content. This blurring of creative authorship may have significant ethical implications in terms of intellectual property, copyright, and the fair compensation of artists and creators. It challenges traditional notions of artistic agency and the rights associated with creative expression, necessitating a reevaluation of legal and ethical frameworks in the age of AI-generated content.

\section*{References}

\medskip

{
\small

[1] Gatys, L. A.,\ Ecker, A. S.,\ \& Bethge, M.\ (2015). A neural algorithm of artistic style.\ {\it arXiv preprint arXiv:1508.06576.}

[2] Radford,\ A.,\ Metz, L.,\ \& Chintala, S.\ (2015). Unsupervised representation learning with deep convolutional generative adversarial networks.\ {\it arXiv preprint arXiv:1511.06434.}

[3] Elgammal, A., Liu, B., Elhoseiny, M., \& Mazzone, M. (2017). Can: Creative adversarial networks, generating" art" by learning about styles and deviating from style norms. {\it arXiv preprint arXiv:1706.07068.}

[4] Hertzmann, A. (2020). Visual indeterminacy in GAN art. {\it In ACM SIGGRAPH 2020 Art Gallery (pp. 424-428).}

[5] Rombach, R., Blattmann, A., Lorenz, D., Esser, P., \& Ommer, B. (2022). High-resolution image synthesis with latent diffusion models. {\it In Proceedings of the IEEE/CVF conference on computer vision and pattern recognition (pp. 10684-10695).}

[6] Kyle Chayka.Is A.I. Art Stealing from Artists?
. {\it https://www.newyorker.com/culture/infinite-scroll/is-ai-art-stealing-from-artists}

[7] Karras, T., Laine, S., Aittala, M., Hellsten, J., Lehtinen, J., \& Aila, T. (2020). Analyzing and improving the image quality of stylegan. {\it In Proceedings of the IEEE/CVF conference on computer vision and pattern recognition (pp. 8110-8119).}

[8] Manovich, L. (2019). Defining AI arts: Three proposals. AI and dialog of cultures" exhibition catalog. {\it Saint-Petersburg: Hermitage Museum.}

[9] Barrat, R. (2018). AI generated nude pictures. {\it https://twitter.com/videodrome/status/983842637525348357}

[10] Klingemann, M. (2017). My Artificial Muse. {\it https://underdestruction.com/2017/06/13/my-artificial-muse/}

[11] Akten, M. (2019). Deep Meditation. {\it https://www.memo.tv/works/deep-meditations/}

[12] Anadol, R. (2019). Latent History. {\it https://refikanadol.com/works-old/latent-history/}

[13] Anadol, R. (2022). Unsupervised — Machine Hallucinations — MoMA. {\it https://refikanadol.com/works-old/unsupervised/}

[14] Allen, J. (2022). Théâtre D’opéra Spatial. {\it https://www.nytimes.com/2022/09/02/technology/ai-artificial-intelligence-artists.html}

[15] Midjourney (2023). Community Showcase (Top). {\it https://www.midjourney.com/showcase/top/}

[16] Qiu, W., \& Legrady, G. (2023, April). Fencing Hallucination: An Interactive Installation for Fencing with AI and Synthesizing Chronophotographs. {\it In Extended Abstracts of the 2023 CHI Conference on Human Factors in Computing Systems (pp. 1-5).}

[17] Klingemann, M. (2019). Hyperdimensional Attractions\: Sirius A. {\it https://www.aiartonline.com/highlights/}
{\it mario-klingemann-3/}

[18] Zhang, L., Rao, A., \& Agrawala, M. (2023). Adding conditional control to text-to-image diffusion models. {\it In Proceedings of the IEEE/CVF International Conference on Computer Vision (pp. 3836-3847).}

[19] Ridler, A. (2019). Mosaic Virus. {\it https://annaridler.com/mosaic-virus }
\section{Supplementary Material}
\newpage
\begin{figure}
    \centering
    \includegraphics[width=1.0\linewidth]{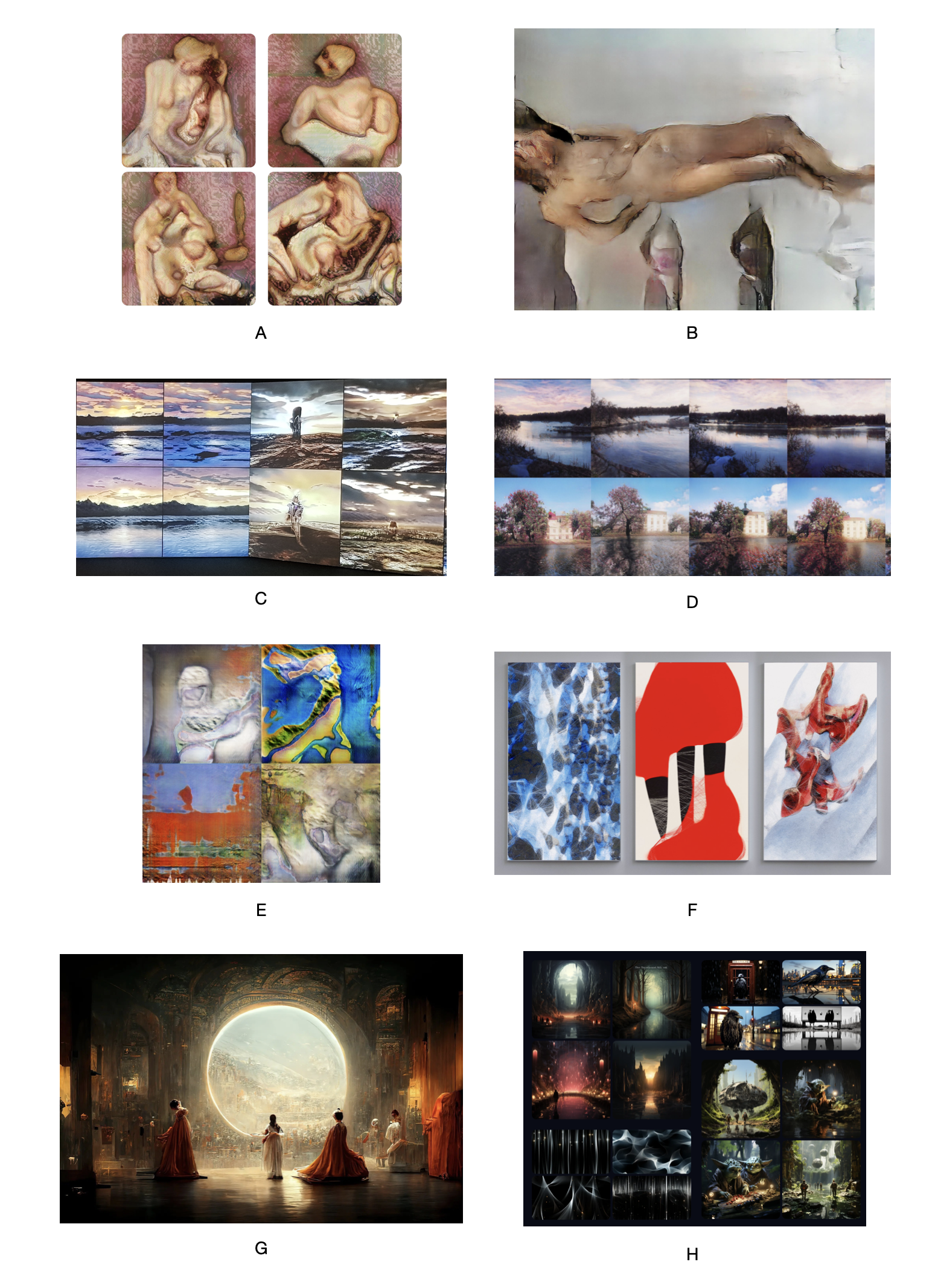}
    \caption{Pairs of AI-generated images of "sameness" from different artists. A \& B: GAN-generated nude photos by Robbie Barrat[9] and Mario Klingemann[10]. C \& D: GAN-generated landscape images by Memo Atken[11] and Refik Anadol[12]. E \& F: GAN-generated abstract art by CAN[3] and [13]. G \& H: Award-winning Midjourney image [14] and the community artworks[15].}
    \label{fig:sameness}
\end{figure}

\begin{figure}
    \centering
    \includegraphics[width=\linewidth]{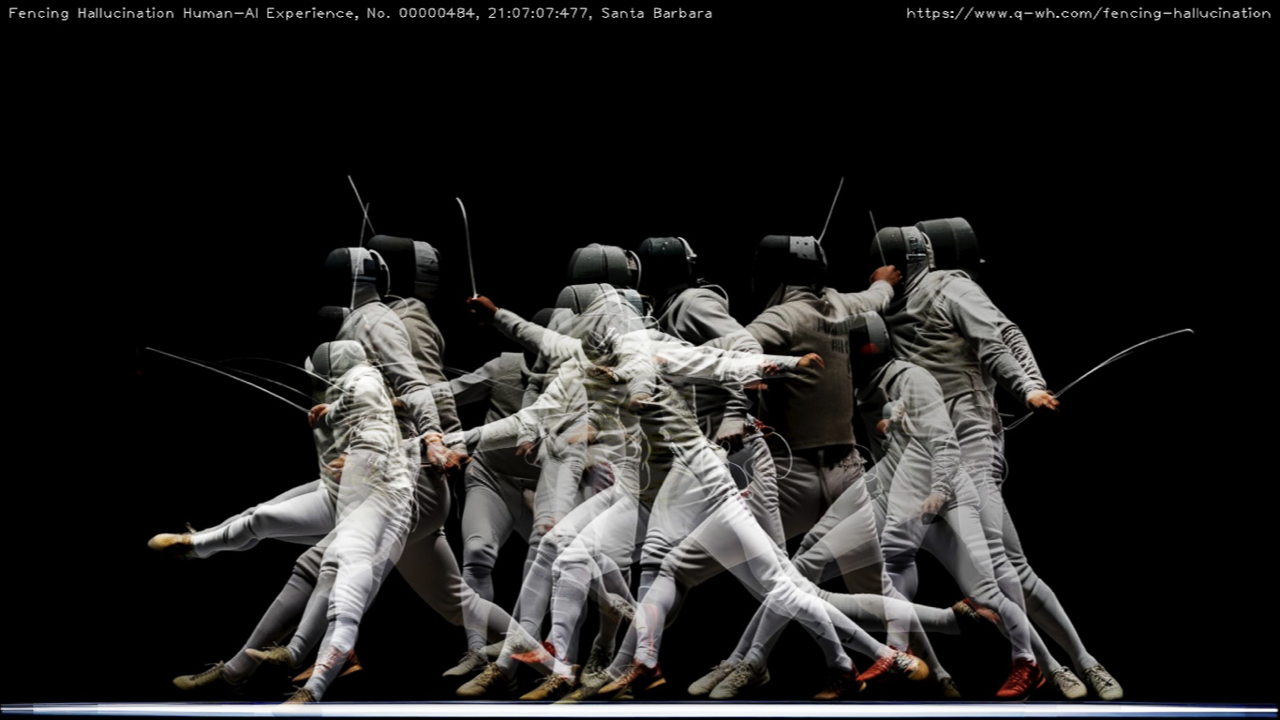}
    \caption{Chronophotograph created by generative AI in Fencing Hallucination.}
    \label{fig:gallery_rep}
\end{figure}

\begin{figure}
    \centering
    \includegraphics[width=\linewidth]{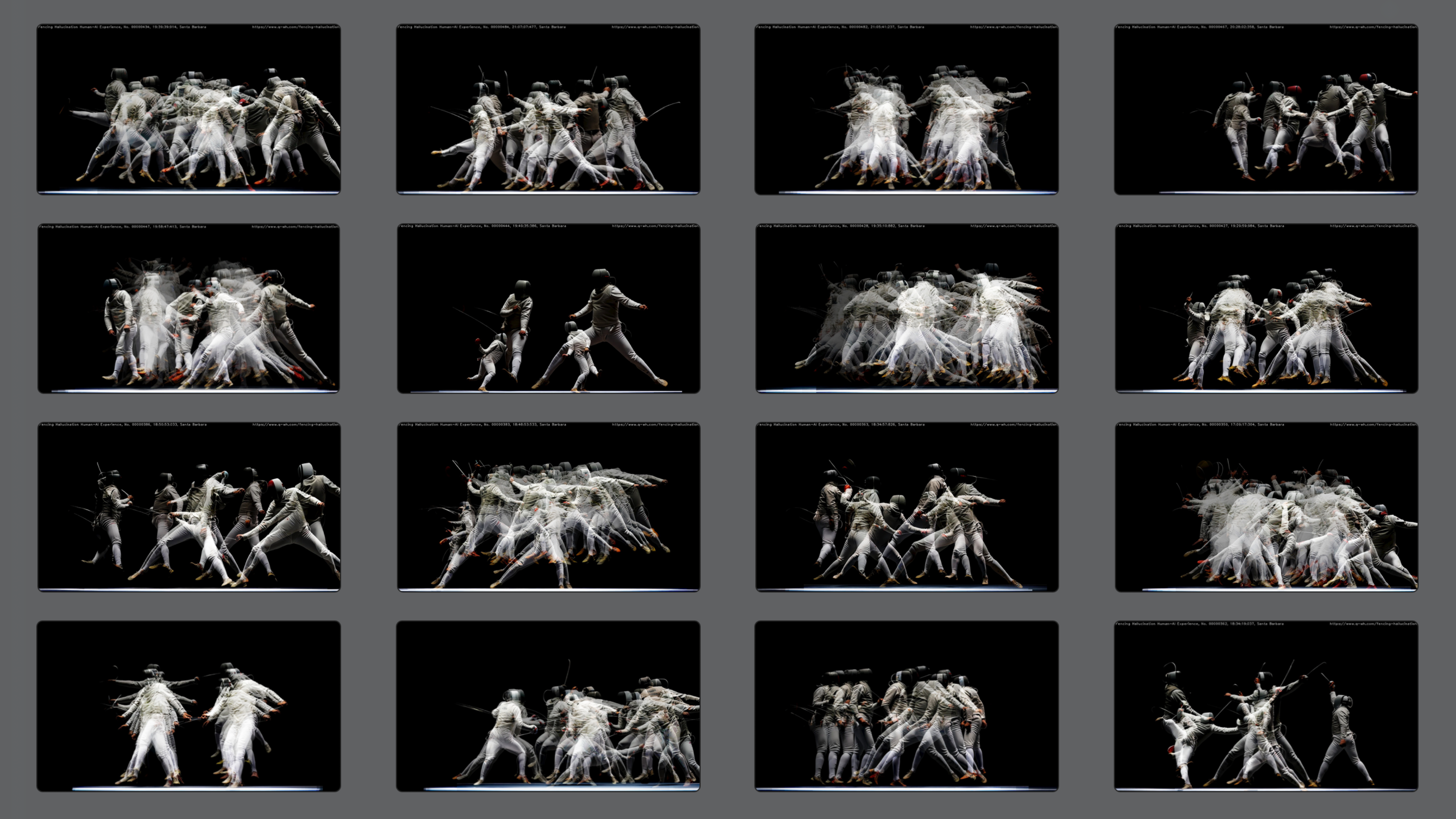}
    \caption{Sampled results from Fencing Hallucination. These images consists of various shapes and patterns, determined by the audience's participation.}
    \label{fig:gallery}
\end{figure}

\begin{figure}
    \centering
    \includegraphics[width=\linewidth]{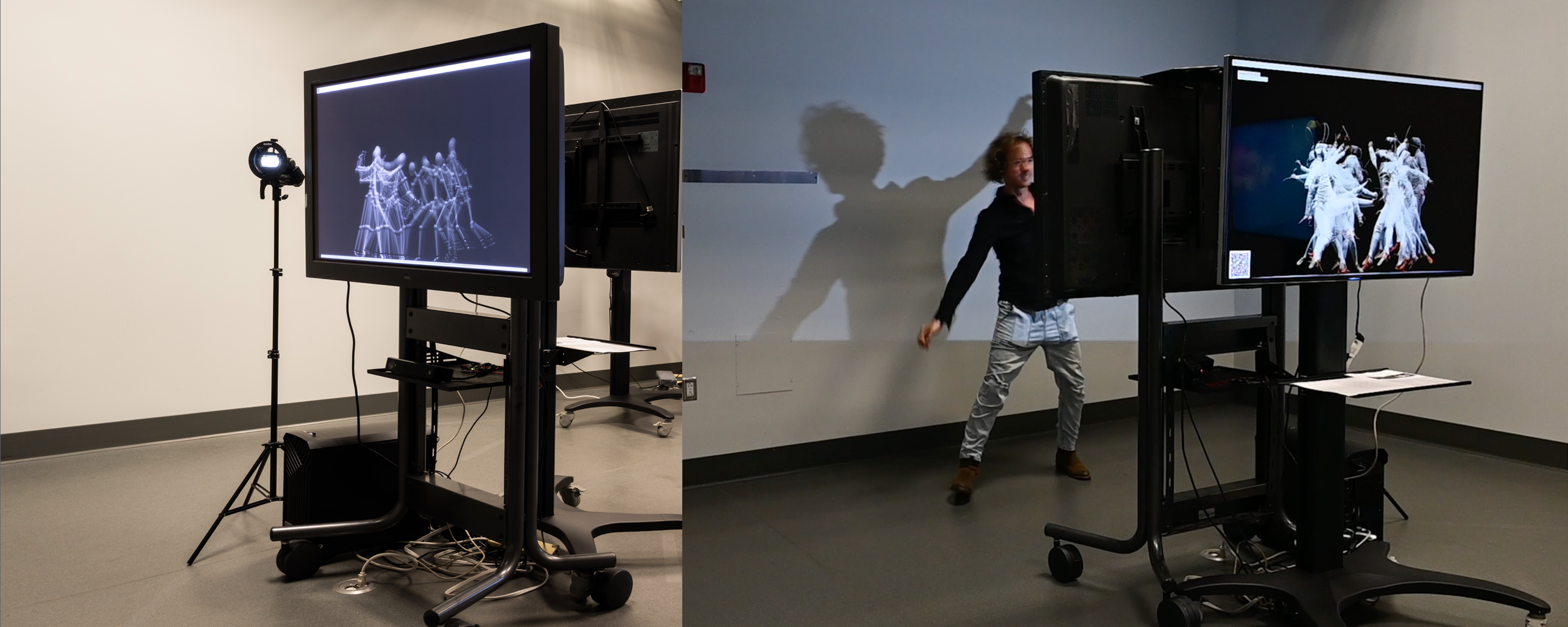}
    \caption{The installation of Fencing Hallucination. Left shows the screen of the real-time content. Right shows the final result of chronophotographs}
    \label{fig:scene}
\end{figure}
\begin{figure}
    \centering
    \includegraphics[width=\linewidth]{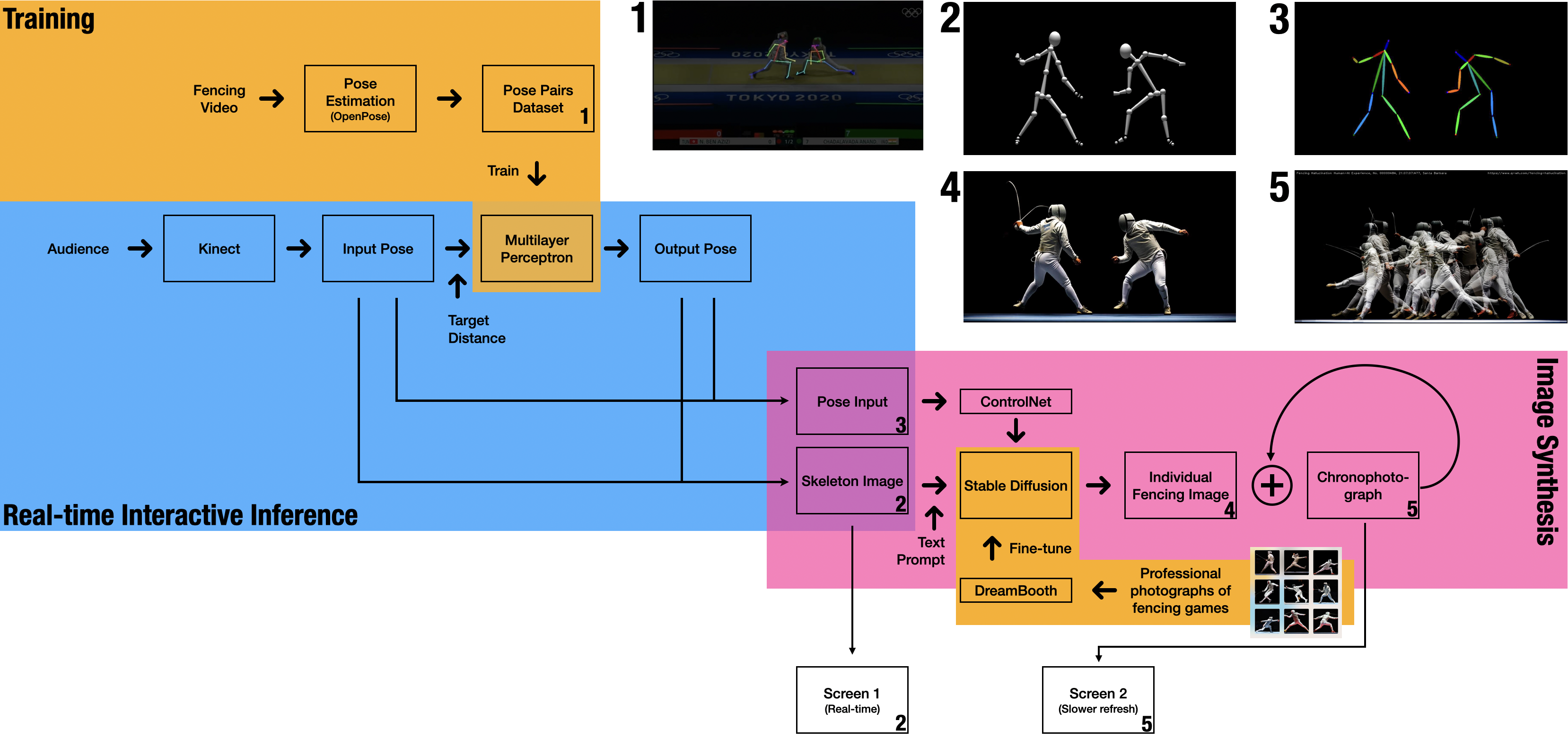}
    \caption{The system workflow description of Fencing Hallucination.}
    \label{fig:workflow}
\end{figure}
\begin{figure}
    \centering
    \includegraphics[width=\linewidth]{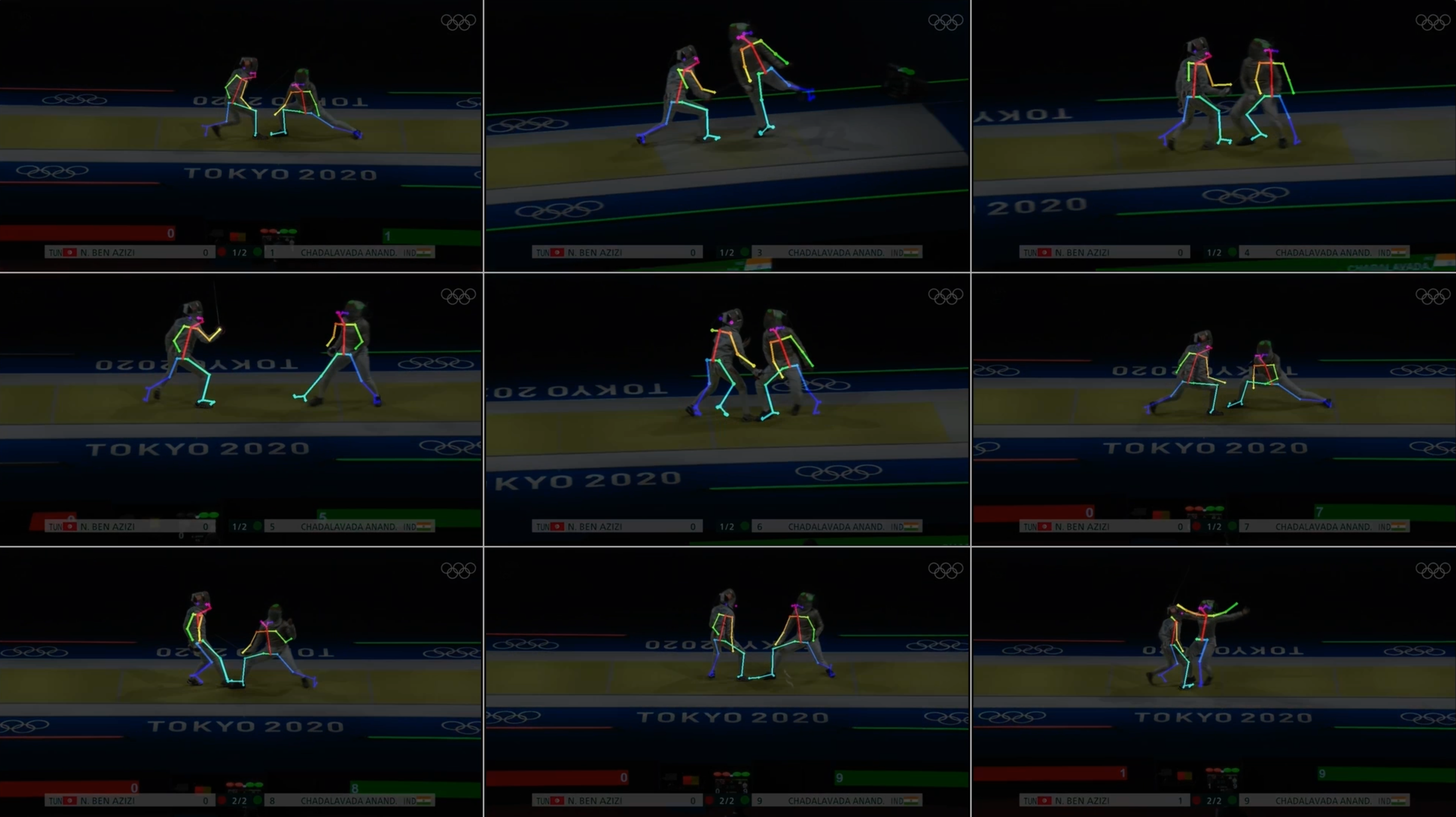}
    \caption{Samples of skeletal data extracted from online fencing videos. These data are collected specially for this project and used to train the real-time AI fencer.}
    \label{fig:dataset}
\end{figure}

\newpage

\end{document}